\newcommand\copyrighttext{%
\footnotesize \textcopyright 2025 IEEE.  Personal use of this material is permitted.  Permission from IEEE must be obtained for all other uses, in any current or future media, including reprinting/republishing this material for advertising or promotional purposes, creating new collective works, for resale or redistribution to servers or lists, or reuse of any copyrighted component of this work in other works.}
\newcommand\copyrightnotice{%
\begin{tikzpicture}[remember picture,overlay]
\node[anchor=south,yshift=10pt] at (current page.south) {\fbox{\parbox{\dimexpr\textwidth-\fboxsep-\fboxrule\relax}{\copyrighttext}}};
\end{tikzpicture}%
}

\documentclass[10pt,twocolumn,letterpaper]{article}

\usepackage{iccv}              

\definecolor{iccvblue}{rgb}{0.21,0.49,0.74}
\usepackage[pagebackref,breaklinks,colorlinks,allcolors=iccvblue]{hyperref}
\usepackage{booktabs} 
\usepackage{multirow} 
\usepackage{array}    
\usepackage{svg}
\usepackage{adjustbox}
\usepackage{enumitem}
\usepackage[accsupp]{axessibility}  
\usepackage{tikz}
\def\paperID{8} 
\def\confName{ICCV}
\def\confYear{2025}

\title{DEAP DIVE: Dataset Investigation with Vision transformers for EEG evaluation}

\author{Annemarie Hoffsommer$^{1}$$^*$, Helen Schneider$^{1}$$^*$, Svetlana Pavlitska$^{1,2}$, J. Marius Zöllner$^{1,2}$\\
\textit{$^{1}$ Karlsruhe Institute of Technology (KIT), Germany}\\
\textit{$^{2}$ FZI Research Center for Information Technology, Germany} \\
{\tt\small helen.schneider@kit.edu}\\
}

\begin{document}
\maketitle
\def\thefootnote{*}\footnotetext{These authors contributed equally to this work}
\copyrightnotice
\thispagestyle{empty}
\pagestyle{empty}

\begin{abstract}
Accurately predicting emotions from brain signals has the potential to achieve goals such as improving mental health, human-computer interaction, and affective computing. Emotion prediction through neural signals offers a promising alternative to traditional methods, such as self-assessment and facial expression analysis, which can be subjective or ambiguous. Measurements of the brain activity via electroencephalogram (EEG) provides a more direct and unbiased data source. However, conducting a full EEG is a complex, resource-intensive process, leading to the rise of low-cost EEG devices with simplified measurement capabilities. 
This work examines how subsets of EEG channels from the DEAP dataset can be used for sufficiently accurate emotion prediction with low-cost EEG devices, rather than fully equipped EEG-measurements.
Using Continuous Wavelet Transformation to convert EEG data into scaleograms, we trained a vision transformer (ViT) model for emotion classification. The model achieved over 91,57\% accuracy in predicting 4 quadrants (high/low per arousal and valence) with only 12 
measuring points (also referred to as channels). Our work shows clearly, that a significant reduction of input channels yields high results compared to state-of-the-art results of 96,9\% with 32 channels. Training scripts to reproduce our code can be found here: \url{https://gitlab.kit.edu/kit/aifb/ATKS/public/AutoSMiLeS/DEAP-DIVE}. 
\end{abstract}    
\section{Introduction}
\label{sec:intro}

The ability to predict human emotions accurately from neural signals holds potential for various fields, including (mental) health\cite{kumar_correlation_2018}, human-computer interaction\cite{zhang_predicting_2021}, and affective computing\cite{kratzwald_deep_2018}. Traditionally, emotion recognition has relied on subjective or ambiguous methods, such as self-assessment and facial expression analysis\cite{she_dive_2021}. While these methods provide valuable insights, they are limited by their dependence on interpretation, leading to inconsistencies and potential biases\cite{chen_understanding_2021}. Importantly, methods such as Electroencephalography (EEG) provide a direct and objective approach to analyzing neural activity and assessing emotional states by measuring the brain's electrical signals. This method mitigates the limitations of traditional techniques, such as subjective biases and inconsistencies.

\begin{figure}
    \centering
    \includegraphics[width=0.7\linewidth, trim={0 0 2cm 0},clip]{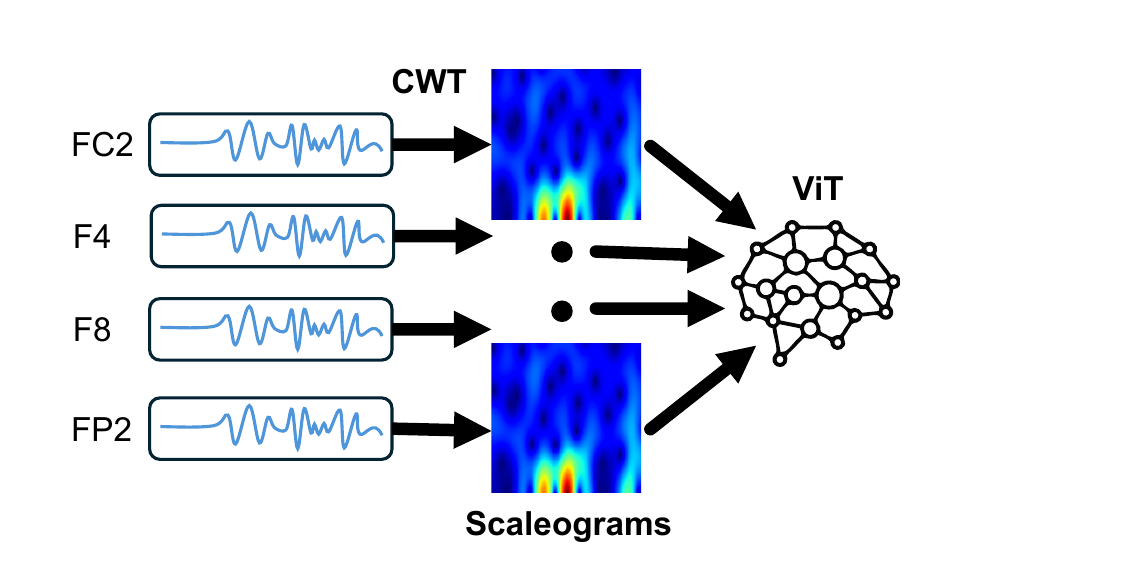}
    \caption{Concept diagram for exemplary approach with 4 channels. Each input consists of only 4 channels with a recording length of one video watched by one person. Each channel is transformed into a separate scaleogram through a CWT. All 4 scaleograms are used as input for the ViT.}
    \label{fig:concept_dir4ch}
\end{figure}

However, despite the advantages of EEG-based emotion prediction, a full-scale EEG setup is often resource-intensive. Medical-grade EEG devices, typically used in clinical settings, require specialized training, costly equipment, and extensive preparation due to the high number of channels involved (up to 60 electrodes). This complexity renders it impractical for application in remote settings, home healthcare environments, regions with limited access to medical resources, or by researchers from different disciplines lacking specialized medical expertise. To address this, there is an increasing interest in rendering portable, low-cost EEG devices usable in practical applications, particularly for emotion recognition \cite{katsigiannis_dreamer_2018}, \cite{dadebayev_eeg-based_2022}. These devices, though limited in terms of the number of electrodes, offer a more accessible solution. An open research challenge involves the use of EEG-devices in delivering reliable predictions of emotions with fewer input channels\cite{jiang_removal_2019}.

This work explores how subsets of EEG channels from the DEAP dataset~\cite{koelstra_deap_2012}, a widely used benchmark for emotion recognition~\cite{tao_eeg-based_2023,bagherzadeh_emotion_2022, Schneider_2025_CVPR}, can be used to achieve accurate emotion prediction, even with low-cost, simplified EEG devices. We investigate the balance between minimizing the number of input channels and maintaining prediction accuracy, with a focus on making EEG-based emotion recognition more accessible and practical in real-world scenarios.

A key element of our approach involves mapping EEG signals to emotions. Emotions are complex, dynamic processes that often defy simple categorization\cite{berrios_what_2019}. However, by employing advanced signal processing techniques, such as Continuous Wavelet Transformation (CWT)\cite{morlet_wave_1982}, which captures temporal dependencies in EEG signals, we can transform these signals into images, known as scaleograms. Therefore, through the transformation of EEG data into CWTs we obtain a visual representation of the DEAP Dataset. Scaleograms are considered meaningful input for data with wavelet characteristics (e.g. EEGs) as they provide direct knowledge to relevant features of the data - the frequencies over time. These images are then used to train a vision transformer \cite{dosovitskiy_image_2021}, a model architecture designed to capture both temporal and spatial dependencies in data. Our approach is displayed visually in an exemplary concept diagram in Figure \ref{fig:concept_dir4ch}.  By pairing CWT with the ViT, we aim to capture the underlying patterns in brain activity that correspond to emotional states. CWT has been used in recent studies combined with convolutional neural networks (CNNs)\cite{garg_emotion_2020}, support vector machines (SVMs) \cite{fairooz_svm_2016}, or standard statistical methods\cite{marjit_eeg-based_2021}.

Our contributions are threefold. \textbf{First}, our work is the first to use CWT combined with vision transformers for emotion recognition and reaching SoTA results.
\textbf{Second}, we offer insights into the minimal requirements for accurate brain-computer interfacing by analyzing different subsets of EEG-channels. We show that using only 12 channels achieves results of 91.57\% accuracy close to SoTA-results of 96.90\% with 32 channels. \textbf{Third}, we are the first, to the best of our knowledge, to provide a baseline for regression on valence and arousal for DEAP with an RMSE of 0.57.
\section{Related Work}
\label{sec:related_work}

\subsection{Emotion Recognition} 
\begin{figure} \centering \includegraphics[width=0.6\linewidth, trim={3cm 3.7cm 3cm 3.7cm},clip]{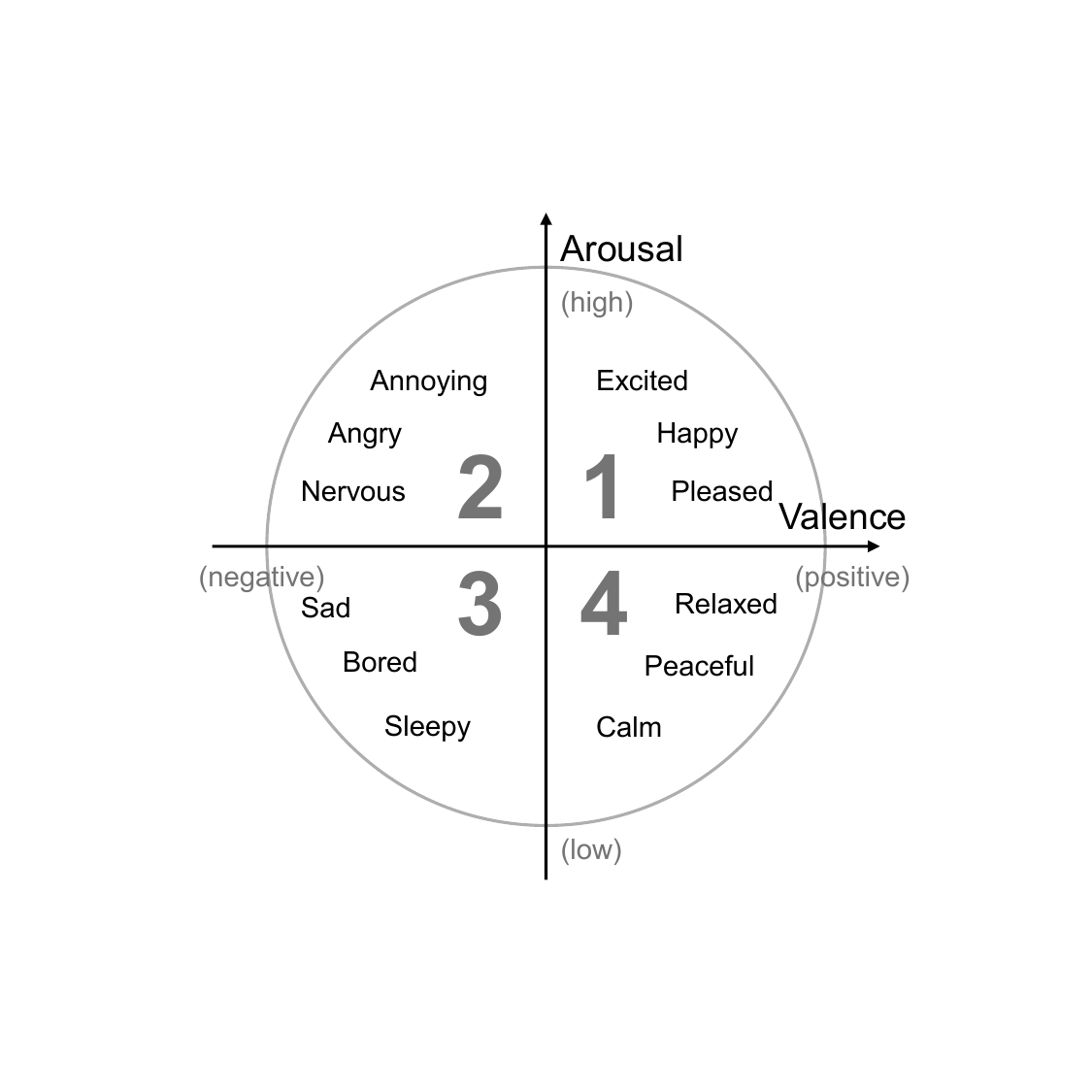} \caption{Russell's Circumplex Model of Affect \cite{dabas_emotion_2018} showing four quadrants(Q) used for our classification.}
\label{fig:emo_s_m} 
\end{figure}

Accurately labeling and classifying emotions, along with understanding their relationships and proximity, is a complex and often ambiguous task, as no clear metric exists. The Circumplex Model of Affect, proposed by Russell in the 1970s \cite{russell_circumplex_1980} and still widely used today, serves as a tool for defining and organizing emotions. This model classifies emotions on a continuous, two-dimensional scale. The horizontal axis -  \textbf{valence} -  represents the range from negative to positive emotions, while the vertical axis - \textbf{arousal} - measures the level of activation, from calmness or sleepiness to excitement or arousal (see Figure \ref{fig:emo_s_m}).

\begin{figure}
    \centering
    \includegraphics[width=0.53\linewidth]{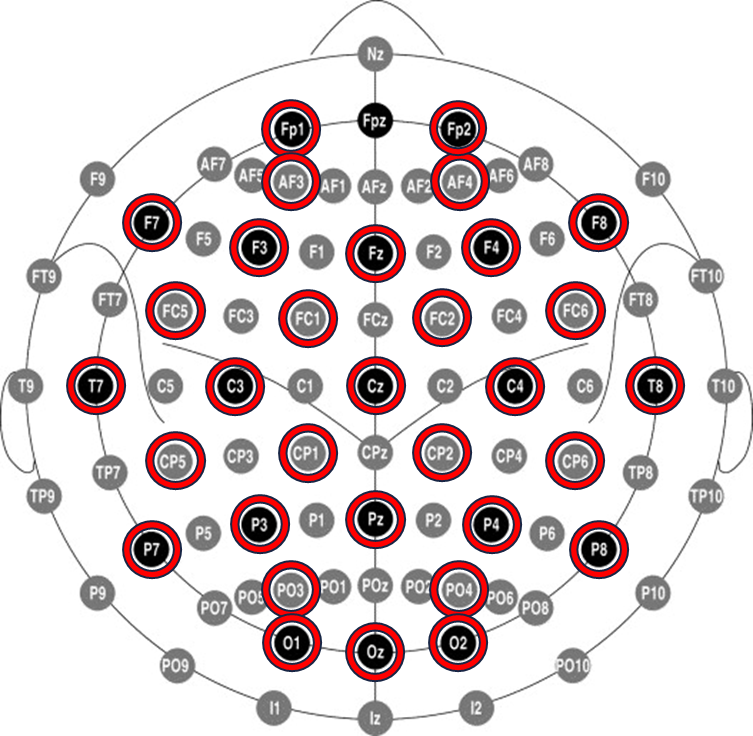}
    \caption{International 10-20-System for EEG-channel placement modified from \cite{oostenveld_five_2001} to highlight channels recorded in DEAP. }
    \label{fig:DEAP Placements}
\end{figure}

Classifying emotions has been done among others with facial configurations \cite{wagner_2024_cage},\cite{sen_continuous_2018}, physiological signals (e.g. heartbeat \cite{shu_wearable_2020_heartbeat}, galvanic skin response\cite{patil_face_2023_gsr}, EEG\cite{fan_eeg_2024}), and body posture \cite{SANTHOSHKUMAR2019_Body}. In our work, we analyze how EEG-data can give insights into what is happening in the brain during emotional occurrences and how new techniques such as vision transformers can help in making low-cost EEG-recordings more useful in affective research.

\begin{table*}[ht]
    \centering
    \caption{Related Work on Emotion Classification on the DEAP Dataset}
    \resizebox{1.0\linewidth}{!}{
    \begin{tabular}{ l c c c c c l c }
    \hline
    \textbf{Author \&} & \textbf{Year} & \textbf{Accuracy $\uparrow$} & \textbf{Accuracy $\uparrow$} & \textbf{Accuracy $\uparrow$} & \textbf{Accuracy $\uparrow$} & \textbf{Method} & \textbf{Number} \\
    \textbf{Citation} & & \textbf{4-Class (\%)} & \textbf{Binary Valence (\%)} & \textbf{Binary Arousal (\%)} & \textbf{Combined Binary (\%)} & &\textbf{Channels} \\
    \hline\hline
    Li et al. \cite{li_emotion_2016} & 2016 & - & 72.06 & 74.12 & 53.41 & CRNN & 32 \\ 
    \hline
    Wang et al. \cite{wang_transformers_2022} & 2020 & 56.93 & 66.51 & 65.75 & 43.73 & Transformer & 32 \\ 
    \hline
    Marjit et al. \cite{marjit_eeg-based_2021} & 2021 & 83.52 & 88.28 & 90.63 & 80.00 & MLP & 32 \\ 
    \hline
    Arjun et al. \cite{arjun_subject_2022} & 2022 & - & 65.90 & 69.50 & 45.80 & LSTM & 32 \\ 
    \hline
    Quan et al. \cite{quan_eeg-based_2023} & 2023 & 69.02 & 81.19 & 79.59 & 64.62 & VAE & 32 \\ 
    \hline
    Tao et al. \cite{tao_eeg-based_2023} & 2023 & - & 97.93 & 97.98 & 95.96 & Attention, CNN & 32 \\ 
    \hline
    Fan et al. \cite{fan_eeg_2024} & 2024 & - & 98.61 & 98.63 & 97.26 & ResNet & 40 \\ 
    \hline
    \textbf{Ours} & & \textbf{91.57} & & & & \textbf{ViT} & \textbf{12} \\ 
    \hline
    \end{tabular}
    }
    \label{tab:related_work_deap_general}
\end{table*}

\begin{table*}[ht]
\setlength{\tabcolsep}{4pt}
    \centering
    \caption{Related Work on Emotion Classification with Wavelet Transformations (WT) on the DEAP Dataset}
    \resizebox{1.0\linewidth}{!}{
    \begin{tabular}{ l c c c l c c }
    \hline
    \textbf{Author \&} & \textbf{Year} & \textbf{Accuracy $\uparrow$} &  \textbf{Accuracy $\uparrow$}  & \textbf{Method} & \textbf{Cross} & \textbf{Number} \\
    \textbf{Citation} & & \textbf{4-Class (\%)}  & \textbf{Binary Combined (\%)} & & \textbf{Subject} &\textbf{Channels} \\
    \hline\hline
    Gupta et al. \cite{gupta_et_al_2019} & 2019 & 72.07 & - & FlexibleWT and Random Forest & yes & 40 \\ 
    \hline
    Islam et al. \cite{Islam_2019_Wavelet} & 2019 & 62.30 & - & DiscreteWT and KNN & no & 10 \\ 
    \hline
    Liu et al. \cite{liu_multi-channel_2020} & 2020 & - & 96.31 & Multi-level features guided Capsule Network & no & 32 \\ 
    \hline
    Garg et al. \cite{garg_emotion_2020} & 2020 & - & 56.44 & Scaleograms and CNN & no & 40\\ 
    \hline
    A.-Conejo et al. \cite{almanza-conejo_emotion_2023} & 2023 & 83.43 & - & Scaleograms and GoogLeNet & no & 32 \\ 
    \hline
    Bagherzadeh et al. \cite{bagherzadeh_emotion_2023} & 2023 & 96.90 & - & Scaleograms and CNN Majority Voting & no & 32 \\ 
    \hline
    \textbf{Ours} & & \textbf{91.57} & & \textbf{Scaleograms and ViT} & & \textbf{12} \\ 
    \hline
    \end{tabular}
    }
    \label{tab:related_work_wavelet}
\end{table*}

\subsection{Datasets for Emotion Recognition using EEG} 
Commonly used datasets for sentiment analysis with EEG data include SEED\&SEED-IV\cite{zheng2015seed1}, \cite{duan2013seed2}, MAHNOB-HCI\cite{soleymani_multimodal_2012} and DEAP \cite{koelstra_deap_2012}. The SEED dataset utilizes movie clips to evoke emotions labeled as positive, negative, and neutral (or happy, sad, fearful, neutral in SEED-IV). However, with only 15 participants, SEED has fewer subjects compared to DEAP, which is seen as a limitation for cross-subject validation. The MAHNOB-HCI dataset, which also uses movie clips to elicit emotions, includes 27 subjects and labels emotions based on liking and disliking. This labeling approach was not chosen for our work due to its misalignment with the intended Circumplex Model of Affect.

The DEAP dataset comprises data from 32 participants, with 40 different measurements (also referred to as channels). These include 32 EEG channels (see Figure \ref{fig:DEAP Placements}), along with additional physiological signals, e.g. electrooculography (eye movement), electromyography (muscle activity), galvanic skin response, blood volume pressure, respiration, and temperature. In DEAP, each participant watched 40 one-minute music videos, selected to evoke a range of emotional responses. After viewing each video, participants rated their emotional state in terms of valence, arousal, dominance, and liking, using a self-assessment method.

For this work, we used the preprocessed DEAP dataset, already prepared via data reordering, downsampling, artifact removal, band-pass filtering, and averaging to a common reference (see \cite{koelstra_deap_2012} for more details).

The number of selected videos in the DEAP Dataset is slightly unbalanced per quadrant (class), while the quadrants are Q1: High Arousal \& High Valence - Q2: High Arousal \& Low Valence - Q3: Low Arousal \& Low Valence - Q4: Low Arousal \& High Valence (see Figure~\ref{fig:emo_s_m}). The following numbers of videos were available in the Dataset with each participant watching all of them: 8 in Q1, 12 in Q2, 10 in Q3 and 10 in Q4. Due to the insignificant low imbalance, we did not employ dataset balancing techniques.

\subsection{State of the art for DEAP dataset classifications}
Previous work for classification on the DEAP dataset can be seen in Table \ref{tab:related_work_deap_general}. The four class accuracy refers to classification of affect into the four quadrants of Q1, Q2, Q3 and Q4 as mentioned above. Therefore predicting valence and arousal simultaneously. Best results are achieved by our vision transformer with 91.57\% accuracy compared to 83.52\% accuracy reached using a Multi-Layer Perceptron (MLP)\cite{marjit_eeg-based_2021}. The binary classifications of valence is predicting either high or low and for the binary classifications of arousal respectively. For comparability to the four class approach, the combined binary accuracy is calculated as the probability of the model for binary valence and the model of binary arousal to classify the same data as a combined model correctly into the four quadrants. The accuracy is calculated by multiplying the accuracy of binary valence and binary arousal. The intention of the binary combined accuracy was solely to provide a theoretical reference for comparison within the 4-class classification task, therefore it was not computed for the proposed model.

\subsection{State of the art DEAP classification using CWT}
Previous work of classifying the DEAP dataset including wavelet transformations for preprocessing are shown in Table \ref{tab:related_work_wavelet}. Using CWT combined with a vision transformer results in 91.57\% accuracy (only 12 channels) reaching near to state-of-the-art results of 96.9\% accuracy (32 channels) with a majority voting of CNNs\cite{bagherzadeh_emotion_2023}, despite significantly lower amount of input and utilizing only one architecture. The highest 4-class accuracy reported by \cite{bagherzadeh_emotion_2023} was achieved using a wavelet transformation. Building on this, the EEG data in this work was represented as scaleograms obtained through a wavelet transformation.

\subsection{State of the art for DEAP dataset regression}
To our knowledge there is no work available that also predicts continuous values from the DEAP dataset. \cite{sen_continuous_2018} predicts the continuous valence of the MAHNOB-HCI dataset by using facial expression and EEG Data. Using a subset of the DEAP Dataset only containing the EEG channels we achieved a Root Mean Squared Error (RMSE) of 0.57.

\section{Methods}
\label{sec:methods}

A CWT was applied to each normalized measurement in order to generate a scaleogram.
The continuous wavelet transformation \cite{morlet_wave_1982}, \cite{grossmann_decomposition_1984}, decomposes a given signal into the frequency components and their magnitude of the signal at each time. This 3D Data can be plotted as a 2D scaleogram with time and frequency as axis and the magnitude as color. Please note that in order to gain a result containing information of time and frequency the CWT looses resolution in difference to a fully reversible Fourier Transformation.

Each resulting image corresponds to one electrode recording from a single participant while watching a one-minute video. The processed images were subsequently input into a vision transformer model. It was possible to choose different subsets of channels and build models using only specific single or groups of measurements. An exemplary approaches 4 channels can be seen in Figure \ref{fig:concept_dir4ch}. 

Although the CWT introduces a slight loss of information into the system, this trade-off was considered acceptable. As the gain in time-domain information, combined with the transformer's ability to process temporal dependencies, was deemed to be beneficial.

 We formulate the emotion recognition problem as a 4-class classification task, with the following labels: high arousal-low valence, high arousal-high valence, low arousal-low valence, low arousal-high valence. The VAQ\_Estimate of the video (Valence Arousal Quadrant) was selected by the DEAP experimenters and was preferred as label over the self assessment manikin (SAM) labels (labeled by the participants) as it is less dependent on differing and inconsistent ratings of persons. If not otherwise stated in Chapter \ref{SAM}, the VAQ\_Estimate was chosen as label.


A vision transformer (Linformer) model was employed.
Vision transformers \cite{dosovitskiy_image_2021} apply the self-attention mechanism of transformers\cite{vaswani_attention_2023}. ViTs partition images into fixed-size patches, linearly embedding them into sequences related to tokens used in language models \cite{dosovitskiy_image_2021}. This approach allows the model to capture long-term dependencies as well as global and spectral context effectively \cite{vaswani_attention_2023}. 

A linear classification layer was added to adapt the network for a 4-class classification task. The target classes were based on high/low arousal and high/low valence. Unless stated otherwise in \ref{SAM}, the VAQ\_Estimate was used as the label for classification.

\section{Experiment Setup}
\label{sec:experiments}

For our experiments, the entire data of the DEAP-dataset was split into k=5 cross validation folds.
The train-test split was conducted not dependent on participant ID. This means that part of the data stemming from a single participant could be in test-split, while the rest of the data from this participant could be in the train-split. This configuration allows for an evaluation of how well the models could generalize when baseline data from a person is available for training. The images obtained by the CWT were bi-linear resized to 224x224 pixels.

For the classification experiments different setups were chosen to evaluate the use of only one channel, subsets of the entire channels and differences between using SAM-labels and VAQ labels. Finally, a regression was conducted in order to have a more fine-grained evaluation of emotional experiences.

For all classification experiments, a learning rate scheduler and the Adam optimizer \cite{kingma_adam_2017} as well as Cross Entropy Loss was employed for the training. Each experiment used 5-fold cross-validation (k=5). Early stopping was implemented, with training halted if the test loss did not improve after five consecutive epochs of exceeding the minimum observed test loss.

\subsection{Classification - Single Channel Experiments}

For the single channel experiments, each EEG channel was tested individually to assess its ability to predict emotions using the smallest available data subset.

\subsection{Classification - Channel Subsets Experiments}

    \begin{figure}
    \centering
    \includegraphics[width=0.5\linewidth]{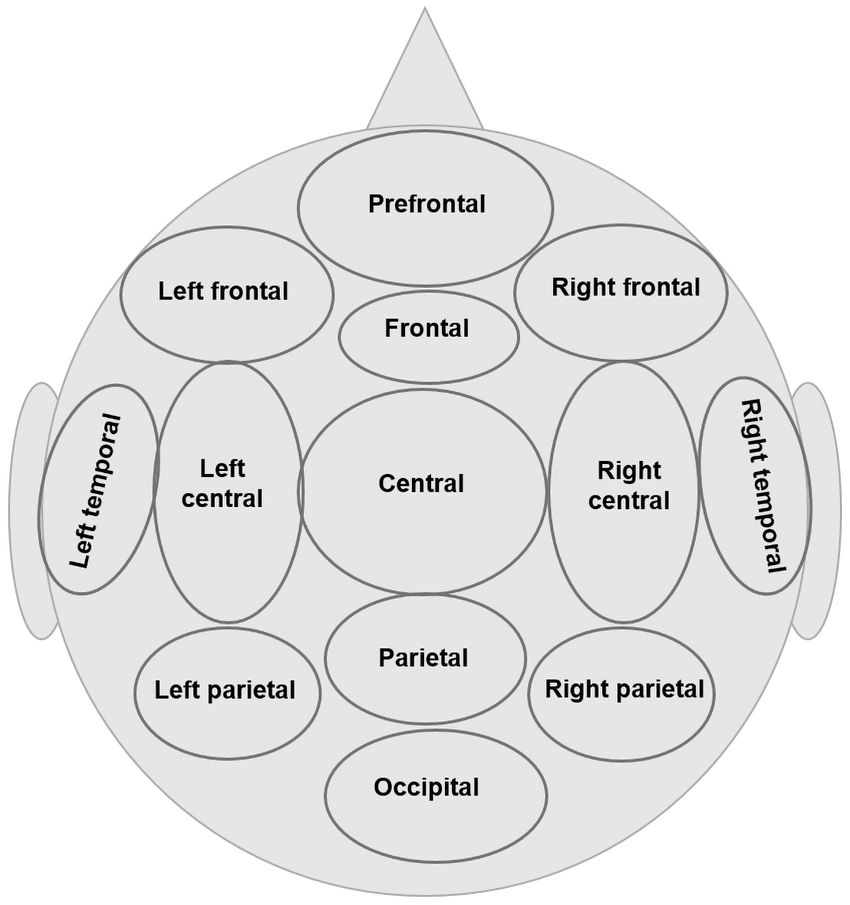}
    \caption{EEG Placement and the corresponding brain region  \cite{maria_volodina_nikolai_smetanin_cortical_2021}.}
    \label{fig:hirnaufbau}
    \end{figure}
    
In order to evaluate how different numbers of EEG-channels result in different accuracies, the following subsets were employed for training:

    \textbf{All Data:} All available data from the DEAP dataset was used for this experiment including EEG and other physiological signals.
    
    \textbf{Muse-12/8/4 Electrodes and Emotiv Testing:} To evaluate the performance of low-budget EEG devices with fewer electrodes, two commonly used devices were selected: the Muse S\footnote{\url{https://choosemuse.com/}}(with 4 electrodes - AF7, AF8, TP9, TP10) and the Emotiv\footnote{\url{https://www.emotiv.com/}}(with 12 electrodes, (AF3, F7, F3, FC5, T7, P7, O1, O2, P8, T8, FC6, F4, F8, AF4)). Since the exact electrode placements of the Muse S are not available in the DEAP dataset, the closest corresponding electrodes were used. The experiments tested the three nearest (setup "Muse-12" - AF3, AF4, FP1, FP2, F7, F8, P7, P8, CP5, CP6, T7, T8), two nearest (setup "Muse-8" - AF3, AF4, F7, F8, P7,P8, T7, T8), and the single nearest (setup "Muse-4") electrodes to match the Muse S setup. As the nearest electrodes to the Muse were unclear to determine there are two Version 4a (AF3, AF4, P7, P8) and 4b (F7, F8, T7, T8). 
    
    \textbf{PCA-12/4 Testing:} 
    A Principal Component Analysis (PCA) was applied to the entire dataset as a standard method to identify the most relevant features. The 12 (setup "PCA-12") or 4 (setup "PCA-4") most relevant channels, determined based on the explained variance ratio, were grouped together and tested.
    
    
    \textbf{EEG-Only Setup:} This experiment used only channels 1-32, which correspond to the EEG electrodes, excluding other physiological measurements.
    
    \textbf{Non-EEG-Only Setup:} In this configuration, only channels 33-40 were used, corresponding to non-EEG physiological measurements.
    
    \textbf{Grouped Channels:} EEG channels were grouped by the brain regions the placements correspond to, and each group’s performance was tested individually. See Figure \ref{fig:hirnaufbau} for the different regions.

\subsection{Classification - Differences between SAM and VAQ labels}
In this setup, the SAM-labels were used instead of the VAQ-labels. SAM labels were given individually by each participant, while the VAQ labels were given by the experimenters. The objective was to investigate potential significant differences between the two labeling methods and gain deeper insights into the limitations of self-assessed data.

For the SAM labels, each video was classified by each participant individually based on the assigned valence and arousal scores of their self-assessment. SAM data ranges from 1 to 9 and was collected by participants selecting a position on a continuous horizontal scale between five manikins, with a distance of 1.6 units between each manikin. To compare the SAM and VAQ label data, the threshold for classifying a video as high or low valence/arousal was set at 5, as in the original DEAP study. 

\subsection{Regression Experiments}
For the regression task only SAM labels were used, as they provide the only continuous values in the dataset. For training, the loss function was adjusted to the Huber Loss to account for outliers that are expected due to human factors and the Root-Mean-Squared-Error(RMSE) was used as metric. As it was observed that regressions tend to train longer, the early stopping threshold was adjusted to 10 consecutive epochs compared to the 5 in the classification settings. 

\section{Results}
\label{sec:results}

Defining what qualifies as relevant, significant, or sufficient for this study does not have a precise answer. Previous works (see Tables~\ref{tab:related_work_deap_general} and~\ref{tab:related_work_wavelet}) do not contain a strict definition of a relevant result. To overcome this problem, we introduce the following new definition: in this work, the results for predicting four classes were considered relevant when the accuracy exceeded twice the expected random accuracy, effectively surpassing the 50\% threshold. We are aware that due to the unbalanced dataset, this is not an exact calculation of chances. Serving the purpose of setting a lower bound, though, this calculation was considered sufficient. 

For regression within the 9x9 area, the expected RMSE for the given data is 3.399. Accordingly, results were deemed relevant when the RMSE was lower than 1.6995, which is half of the expected RMSE value.

\subsection{Classification - Single Channel Experiments}

\begin{figure*}[ht]
  \centering
  \adjustbox{trim=0cm 0.5cm 0cm 0cm, clip}{%
   \includegraphics[width=1\linewidth]{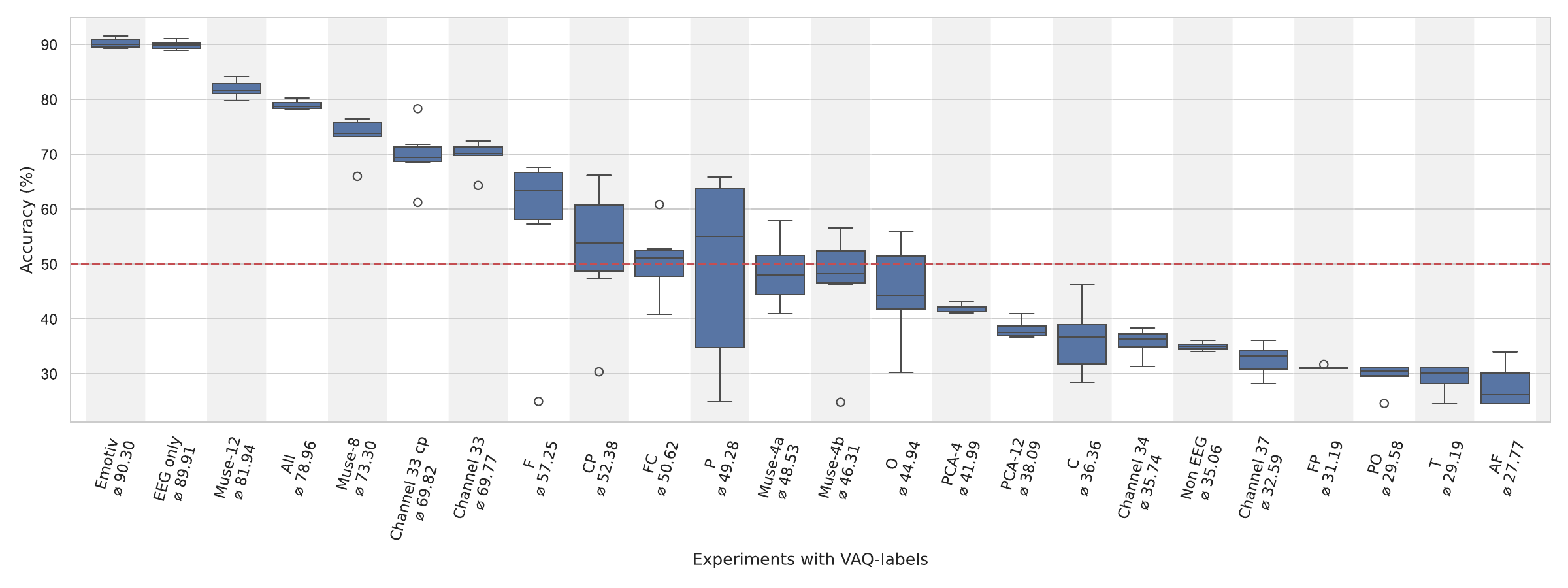}
    }
  \caption{Classification experiments with VAQ labels - labels given by the experimenters. Results of Table~\ref{tab:Ungrouped_singles},~\ref{tab:ungrouped SAM} and~\ref{tab:ungrouped_experiments_Buchstaben} are presented visually by their 5 folds locality, spread and skewness through their quartiles. Experiments are ordered by descending mean of their 5 folds accuracy($\uparrow$) respectively. The mean (Ø) is shown after the experiment name in the labels of the horizontal axis. The dashed red line indicates the threshold of the double of random results. The data shows the first 6 experiments clearly over the threshold line in all 5 folds respectively. Outliers that differ significantly from the other folds are depicted as circles. Interestingly, the grouped "O"-channels perform only slightly below the threshold with a mean of 44.94\% even though the group only contains 3 channels. The Emotiv-channels perform best with a mean of 90,3\% in accuracy with only 12 channels as input. }
  \label{fig:boxplot_all_nosam}
\end{figure*}

\begin{table}[t]
\centering
\setlength{\tabcolsep}{4pt} %
\caption{Accuracy ($\uparrow$) for single experiments}
\label{tab:Ungrouped_singles}
\resizebox{1.0\linewidth}{!}{
\begin{tabular}{|c|c|c|c|c|c|c|}
\hline
\textbf{Channels} & \textbf{Fold 1} & \textbf{Fold 2} & \textbf{Fold 3} & \textbf{Fold 4} & \textbf{Fold 5} & \textbf{Mean} \\
\hline
channel 33 & 70.38 & 64.36 & 70.00 & \textbf{72.44} & 71.67 & 69.37 \\ \hline
channel 34 & 37.05 & 38.33 & 31.41 & 34.62 & 37.31 & 35.74 \\ \hline
channel 37 & 28.21 & 30.26 & 36.15 & 33.97 & 34.36 & 32.19 \\ \hline
channel 33 & 71.79 & 68.57 & 61.25 & 69.17 & \textbf{78.33} & 69.82 \\
cross person &&&&&&\\ 
\hline
\end{tabular}
}
\end{table}
Only one channel, Channel 33 produced a sufficiently high classification accuracy when used as the sole input. This channel, representing horizontal eye movement (hEOG), is not typically part of an EEG setup. With an accuracy exceeding 70\% for classifying four classes, it provides a partial answer to the question of how little input is needed to achieve meaningful predictions. Other channels did not yield an accuracy considered as significant.

As follow up we conducted a cross-person experiment for the single channel of horizontal eye movement. In this setup the test and train split was depending on the persons. Meaning that no data from a person selected in test was used for training of the model. The accuracy remained the same for the cross-person setting of channel 33, one fold even surpassed the previous experiments. This suggests that the eye movement could be a beneficial measurement for low-budget EEG-devices as it provides a solid basis for classification. However, it is yet to be determined whether the eye-movement-model predicted an emotion or rather just learned which eye movements correspond to which videos and what label the video corresponded to.
For detailed accuracy of each fold of the best single channel results, see Table~\ref{tab:Ungrouped_singles}. 
A box-plot overview of all conducted experiments (including the best four single channel results) and their means over all 5 folds can be seen in Figure~\ref{fig:boxplot_all_nosam}. Channel 33 clearly lies in the top 7 results of all experiments.

For completeness, we experimented with cross-person setting by leaving 6 people out for each fold for validation. For channel 33 this resulted in a mean accuracy of 71.9\%.

\subsection{Classification - Channel Subsets Experiments}

    \textbf{All Data -} When all available channels were utilized, the classification accuracy reached approximately 80\%.
    
    \textbf{Muse-12/8/4 Electrodes -} Using 12 electrodes resulted in one of the best-performing configurations with a maximum accuracy of 84\%. Reducing the input to 8 electrodes caused an 8\% drop in accuracy, reaching 76\%. However, further reducing the input to 4 electrodes led to a more substantial drop in accuracy, falling by approximately 20\% to 58-56\%. This indicates that the model's performance is more resilient when reducing from 12 to 8 electrodes than when reducing from 8 to 4 electrodes. See Table~\ref{tab:ungrouped_experiments} for comparison of accuracy across all folds and the corresponding number of channels for each channel-name.
    \begin{figure}[htbp]
    \centering
    \adjustbox{trim=0cm 0cm 0cm 0.27cm, clip}{%
        \includegraphics[width=1\linewidth]{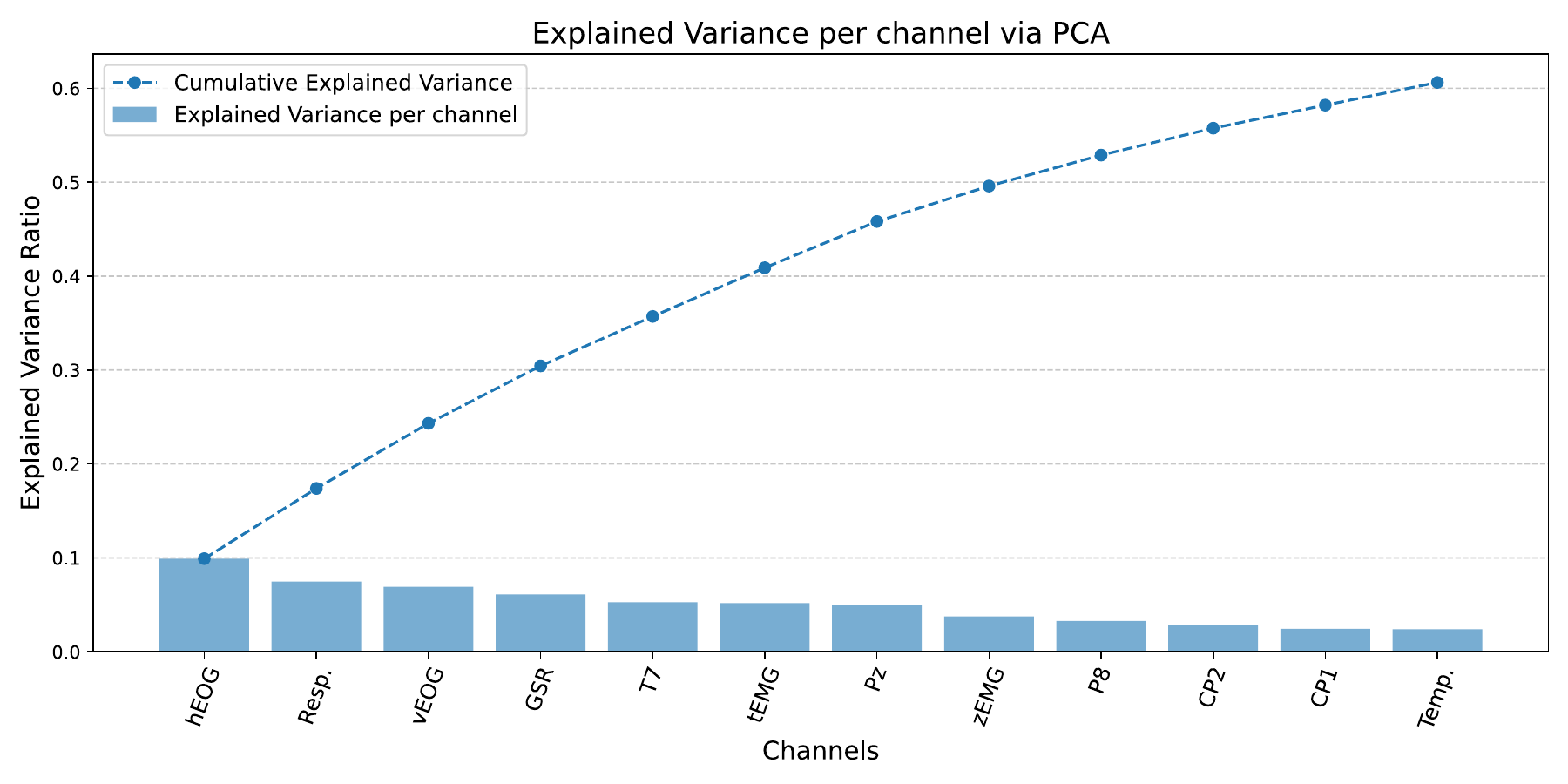}
    }
    \caption{Explained variance for the best 12 channels obtained through PCA, with channels sorted in descending order of explained variance. The line shows the cumulative explained variance across the sorted channels. The most relevant features according to the explained variance is the hEOG (channel 33). It can be seen that the 12 channels only cumulate in a explained variance of 60\%; more than half of the channels are needed to obtain a cumulative explained variance above the 80\% mark.}
    \label{fig:pca}
    \end{figure}
    \textbf{Emotiv -} The channels corresponding to those used in the Emotiv device achieved the best classification accuracy at 91.50\%. This suggests that this specific channel subset performs exceptionally well, even with significantly fewer (12 of 40) channels than the full dataset and shows one of the best trade offs between accuracy and used channels.

    \textbf{PCA -} The explained variance for each channel obtained through PCA is shown in Figure~\ref{fig:pca}. The ordering of explained variance across channels does not align with the ranking obtained from single-channel testing. However, the highest-ranked channel (hEOG - channel 33) is consistent between both methods. Grouping the top 12 channels resulted in a maximum accuracy of 41.04\%, while grouping the top 4 led to a slightly better performance with 43.16\%. Consequently, no configuration based on PCA findings is considered relevant. These results are explainable since PCA is a linear feature reduction method~\cite{anowar_conceptual_2021}, while the relationships between the features are not necessarily linear.

    \textbf{EEG-only and Non-EEG only -} The EEG-only configuration achieved 91.06\% accuracy, ranking second-best with no significant difference to the top result. However, this setting required more than double the number of channels compared to the Emotiv configuration for a similar outcome. In contrast, the non-EEG channel setup did not produce sufficient accuracy, despite containing channel 33.
    
    \textbf{Grouped Channels -}
    Grouping the channels by location of the electrode led to ambivalent results, shown in Table~\ref{tab:ungrouped_experiments_Buchstaben}. The groups frontal (f), parietal (p), and  between central and parietal (cp) showed at least one fold with sufficient accuracy. But there is at least one fold per group where no generalization is developed. 
    This might show that EEG features are less obvious, which might lead to an outlier-prone prediction when less training data is available. 
    Furthermore, there are differences in the number of channels that are part of each group. All groups with relevant results are on the upper end of channel numbers, whereas the channels containing fewer channels tend to be less performing. For example, the comparison of group fc and cp shows that a group with more channels does not directly yield higher accuracy but seems more dependent on the input. 
    Training data limitations could be a reason for more stable training with more channels; however, as the single-channel experiment showed, the used architecture is capable of performing a stable classification with only one channel.  
    

    See Figure~\ref{fig:boxplot_all_nosam} for a better insight into the experiments and a visual direct comparison between the experiments.

\begin{table}[t]
\centering
\setlength{\tabcolsep}{4pt} %
\caption{Accuracy ($\uparrow$) for experiments, VAQ labels}
\label{tab:ungrouped_experiments}
\resizebox{1.0\linewidth}{!}{
\begin{tabular}{|c|c|c|c|c|c|c|c|}
\hline
\textbf{Channels} & \textbf{\#} & \textbf{Fold 1} & \textbf{Fold 2} & \textbf{Fold 3} & \textbf{Fold 4} & \textbf{Fold 5} & \textbf{Mean} \\
\hline
all & 40 & 78.12 & 78.44 & 78.39 & \textbf{80.23} & 79.61 & 78.96 \\ 
\hline
eeg-only & 32 & 88.94 & 89.18 & \textbf{91.06} & 90.37 & 89.99 & 89.91 \\
\hline
non-eeg & 8 & 35.51 & 36.13 & 35.09 & 34.10 & 34.48 & 35.06 \\
\hline
Emotiv & 12 & 89.70 & 89.56 & \textbf{91.57} & 91.28 & 89.37 & 90.30 \\ 
\hline
Muse-12 & 12 & 83.25 & 81.33 & \textbf{84.25} & 79.77 & 81.10 & 81.94 \\
\hline
Muse-8 & 8 & 73.35 & 74.30 & \textbf{76.44} & 66.01 & 76.38 & 73.30 \\
\hline
Muse-4a & 4 & 40.97 & 52.62 & 43.50 & 58.06 & 47.48 & 48.53 \\
\hline
Muse-4b & 4 & 47.23 & 53.45 & 56.65 & 24.85 & 49.37 & 46.31 \\
\hline
PCA-12 &12&38.90&36.88&41.04&36.98&36.67&38.01\\
\hline
PCA-4&4&41.09&42.23&43.16&42.28&41.21&41.99\\ \hline
\end{tabular}
}
\end{table}

\begin{table}[h!]
\centering
\setlength{\tabcolsep}{4pt} 
\caption{Accuracy ($\uparrow$) for experiments, grouped by electrode placement, VAQ labels}
\label{tab:ungrouped_experiments_Buchstaben}
\resizebox{1.0\linewidth}{!}{
\begin{tabular}{|c|c|c|c|c|c|c|c|}
\hline
\textbf{Channels} & \textbf{\#} & \textbf{Fold 1} & \textbf{Fold 2} & \textbf{Fold 3} & \textbf{Fold 4} & \textbf{Fold 5} & \textbf{Mean} \\
\hline
t  & 2 & 24.62 & 27.89 & 31.15 & 31.15 & 31.15 & 29.19 \\ \hline
f  & 5 & 25.00 & 66.88 & \textbf{67.66} & 66.02 & 60.70 & 57.25 \\ \hline
c  & 3 & 28.51 & 46.33 & 37.11 & 30.26 & 39.58 & 36.36 \\ \hline
fp & 2 & 30.96 & 30.96 & 31.73 & 31.15 & 31.15 & 31.19 \\ \hline
af & 2 & 30.96 & 34.04 & 24.62 & 24.62 & 24.62 & 27.77 \\ \hline
po & 2 & 31.15 & 29.81 & 31.15 & 31.15 & 24.62 & 29.19 \\ \hline
fc & 5 & 40.92 & 52.82 & 51.65 & 60.87 & 46.85 & 50.62 \\ \hline
cp & 4 & 55.29 & 30.39 & 47.48 & \textbf{66.17} & 62.57 & 52.38 \\ \hline
o  & 3 & 56.01 & 53.60 & 41.14 & 30.26 & 43.67 & 44.94 \\ \hline
p  & 5 & \textbf{65.86} & 64.84 & 60.78 & 24.92 & 30.00 & 49.28 \\ \hline
\end{tabular}
}
\end{table}

\subsection{Classification - Differences between SAM and VAQ labels}
\label{SAM}
Using the SAM labels resulted in mostly comparable accuracy to the VAQ label results. Taking into account that the SAM labels are more dependent on each participant and therefore have a more volatile influence on the results, the results for all channels, EEG-only, Muse-12 and Muse-8 are viewed as similar to the results of the VAQ labels. In the Emotiv and Muse-4b setup there is a similar tendency of fold-performance in the SAM results and the VAQ results, respectively. However the overall difference of the Emotiv-Sam and Emotiv-VAQ is much larger than the overall difference of the Muse-4b-SAM and Muse-4b-VAQ. For the Muse-4b setup an explanation could be, that the training already seems to be unstable in the original setup (see fold 4 of Muse-4b in Table~\ref{tab:ungrouped_experiments}) and the differences and uncertainties of the labeling persons are disadvantageous. This is even more reasonable as 671 out of the 1280 total scaleograms were grouped into a different quadrant by the SAM labels than by the given VAQ.

Channel 33 showed no generalization in this setup.
It is possible, that the horizontal eye movement (channel 33) is corresponding to the dynamics of the presented videos more than to the experienced feelings. The drastic differing of results could support this hypothesis, if the SAM labels are assumed to be more accurate to the experienced emotions. However eye movement is used to predict emotional states in~\cite{hu_exploration_2021} ,\cite{tarnowski_eye-tracking_2020} with high accuracy. The open question whether the experienced emotions or the video dynamics are crucial for the prediction is also relevant for the EEG channels. An interesting result regarding this question is the performance of the Occipital Lobe group 'o' in Table~\ref{tab:ungrouped_experiments_Buchstaben} with 56\% accuracy for 3 channels. As the Occipital Lobe is associated as center of processing visual input~\cite{javed_neuroanatomy_2024}.

\begin{figure}[t]
  \centering
  \adjustbox{trim=0cm 0.5cm 0cm 0cm, clip}{%
    \includegraphics[width=1\linewidth]{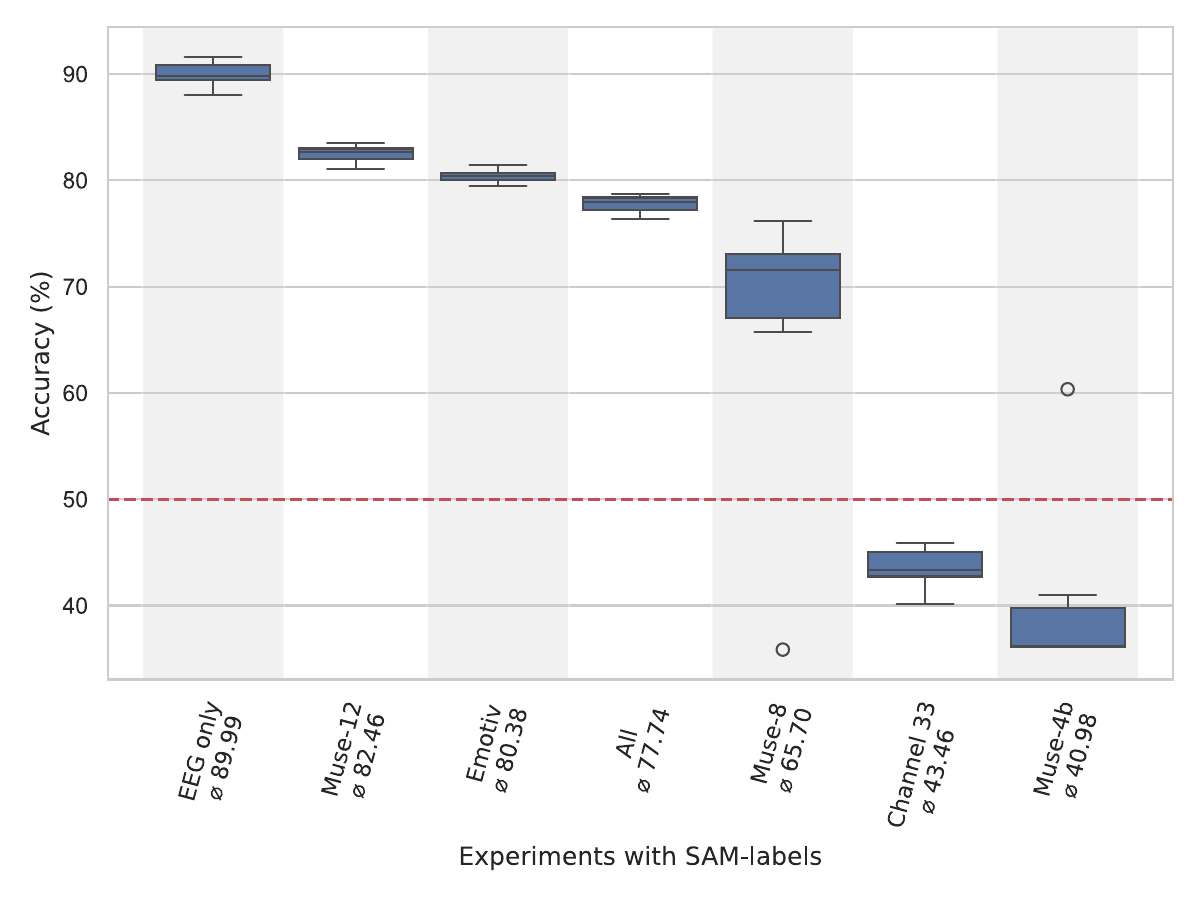}
  }
  \caption{Box-plot and mean of the classification experiments with self-assessment labels (SAM) - labels given by participants. Results of Table~\ref{tab:ungrouped SAM} are presented visually as box-plots of their 5 folds and ordered by descending mean accuracy($\uparrow$). The mean (Ø) is shown after the experiment name in the labels of the horizontal axis. The dashed red line indicates the threshold of the double of random results. In contrast to the VAQ-label experiments seen in Figure~\ref{fig:boxplot_all_nosam} only four experiments clearly exceed the threshold line with all 5 folds. }
  \label{fig:boxplot_sam}
\end{figure}

Detailed results are shown in Table~\ref{tab:ungrouped SAM}, a box-plot overview is provided in Figure~\ref{fig:boxplot_sam}.

\begin{table}[t]
\centering
\setlength{\tabcolsep}{4pt} 
\caption{Accuracy($\uparrow$) with SAM labels, Differences between the maximal Accuracies from SAM labels to VAQ labels}
\resizebox{1.0\linewidth}{!}{
\begin{tabular}{|r|c|c|c|c|c|c|c|}
\hline
\textbf{Channels}& \textbf{\#} & \textbf{Fold 1} & \textbf{Fold 2} & \textbf{Fold 3} & \textbf{Fold 4} & \textbf{Fold 5}&\textbf{Diff.} \\ \hline
all      &  40    & 76.33 & 77.04 & 78.44 & 78.69 & 78.19& -1.84 \\ \hline
eeg-only &32& 88.06 & 91.18 & 89.68 & 89.40 &91.63 & +0.57 \\ \hline
Emotiv &12& 80.71 & 79.87 & 80.45 & 81.40 & 79.48& -10,17 \\ \hline
Muse-12   &12& 83.05 & 82.82 & 81.10  & 81.82 & 83.51&-0.74 \\ \hline
Muse-8    & 8&70.91 & 35.85 & 73.28 & 76.18 & 72.29&-0.26 \\ \hline
Muse-4b   & 4&36.12 & 36.12 & 36.12& 60.34 & 36.21&+3.69 \\ \hline
Channel 33   &1& 40.13 & 43.21& 45.90 & 45.51 & 42.56&-26.54 \\ \hline
\end{tabular}
}
\label{tab:ungrouped SAM}

\end{table}

\subsection{Regression Experiments}
In the regression experiments, the EEG-only setup produced the best performance, achieving the lowest RMSE of 0.57. Unlike the classification task, the Emotiv setup did not perform as well, reaching an RMSE of 0.88, which is still the second-best result. 

The experiments with Muse-12 and Muse-8 showed similar tendencies comparing to the classification results, with a small performance gap between them. As observed in the classification task, the performance drop when reducing to 4 channels was more substantial, indicating a larger disparity in accuracy when fewer channels were used.

For channel 33 the results were not considered relevant but are still noticeably better than random. Here it should be taken into account that one channel only provides a limited amount of data for training.

\begin{figure}[t]
  \centering
  \adjustbox{trim=0cm 0.5cm 0cm 0cm, clip}{%
    \includegraphics[width=1\linewidth]{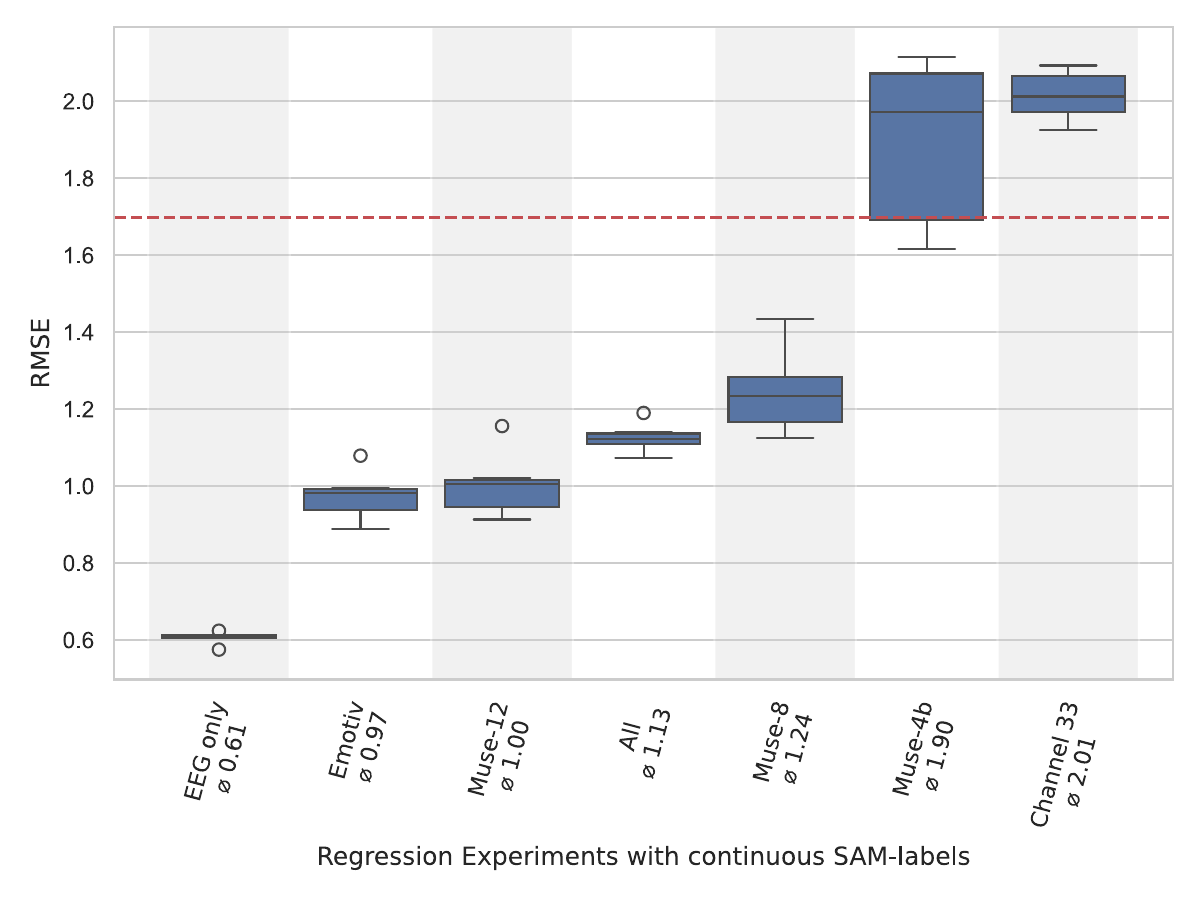}
  }
  \caption{Box-plot and mean RMSE of the best seven regression experiments with continuous SAM labels. Results of Table~\ref{tab:regression_ungrouped} are presented visually as box-plots of their 5 folds and ordered by ascending mean RMSE($\downarrow$). The EEG-only and Emotiv experiments show best results, with Muse-12 close to the results of Emotiv.}
  \label{fig:regression}
\end{figure}

A detailed list of results is shown in Table~\ref{tab:regression_ungrouped}, a box-plot overview is provided in Figure~\ref{fig:regression}.

\begin{table}[t]
\centering
\setlength{\tabcolsep}{4pt} 
\caption{RMSE Values ($\downarrow$) for Each Fold, Regression}
\label{tab:regression_ungrouped}
\resizebox{1.0\linewidth}{!}{
\begin{tabular}{|r|c|c|c|c|c|c|c|}
\hline
\textbf{Channels} & \textbf{\#} & \textbf{Fold 1} & \textbf{Fold 2} & \textbf{Fold 3} & \textbf{Fold 4} & \textbf{Fold 5} & \textbf{Mean} \\
\hline
all        & 40 & 1.074  & 1.140  & 1.119  & 1.190  & 1.107  & 1.126 \\ \hline
eeg-only   & 32 & \textbf{0.575} & 0.611  & 0.624  & 0.612  & 0.604  & 0.605 \\ \hline
Muse-12    & 12 & 1.156  & 0.926  & 1.020  & \textbf{0.913}  & 1.005  & 1.004 \\ \hline
Muse-8     & 8  & 1.124  & 1.147  & 1.296  & 1.434  & 1.223  & 1.245 \\ \hline
Muse-4b    & 4  & 2.081  & 1.615  & 2.116  & 2.046  & 1.624  & 1.896 \\ \hline
Emotiv     & 12 & 0.994  & 1.079  & 0.989  & \textbf{0.889}  & 0.924  & 0.975 \\ \hline
Channel 33 & 1  & 1.926  & 2.011  & 2.093  & 1.958  & 2.082  & 2.014 \\ \hline
\end{tabular}
}
\end{table}

\section{Conclusion}
\label{sec:conclusion}

We conducted a classification and regression with different subsets of the DEAP dataset, up to testing only single input EEG channels. In conclusion, our research underscores the effectiveness of pairing scaleograms and vision transformer as well as predicting emotional states with only subsets of a common EEG.

We evaluated the trade-off between minimal input and prediction performance, evaluating models trained on varying numbers of channels. Through these analyses, we showed different configurations that can be used for portable EEG devices, enabling more accessible, reliable emotion prediction technologies, while also offering significant contributions to the broader research community.

The performance of our model showed some ambiguities with different experiment setups.    
Pairing subsets that achieved a high accuracy with additional channels does not always lead to better performances. Most visible was this phenomenon in experiments where all available inputs were used but did not lead to best model performance. The underlying mechanisms of this non-superpositional behavior and the factors contributing to the amplification of accuracy across certain channels remain open questions for further investigation in the field of explainable AI. An interdisciplinary approach, particularly one integrating insights from neuroscience, will be highly beneficial in advancing our understanding of these phenomena.


\section*{Acknowledgment}
This work was supported by funding from the Topic Knowledge for Action of the Helmholtz Association (HGF).
{
    \small
    \bibliographystyle{ieeenat_fullname}
    \bibliography{references_zotero, egbib}
}

\end{document}